\pdfoutput=1
\documentclass[11pt]{article}
\usepackage{emnlp2021}
\usepackage{times}
\usepackage{latexsym}
\usepackage[T1]{fontenc}
\usepackage[utf8]{inputenc}
\usepackage{microtype}
\usepackage{amsmath}
\usepackage{amsthm}
\usepackage{amssymb,amsfonts} 
\usepackage{algorithm}
\usepackage{algorithmic}
\usepackage{booktabs}
\usepackage{graphicx}
\usepackage{url}
\usepackage{textcomp}
\usepackage{xcolor}
\usepackage{subfigure} 
\usepackage{float}
\usepackage{multirow}
\usepackage{multicol}
\usepackage{color}
\usepackage{bm}
\usepackage{bbm}
\usepackage{enumitem}
\usepackage[normalem]{ulem}

\usepackage{pifont}

\def \R{\mathbb R}

\newcommand{\bx}{\mathbf{x}}
\newcommand{\bs}{\mathbf{s}}
\newcommand{\by}{\mathbf{y}}

\newcommand{\bX}{\mathbf{X}}
\newcommand{\bY}{\mathbf{Y}}
\newcommand{\bH}{\mathbf{H}}
\newcommand{\bW}{\mathbf{W}}
\newcommand{\bI}{\mathbf{I}}

\newcommand{\bD}{\mathbf{D}}
\newcommand{\bA}{\mathbf{A}}

\newcommand{\cG}{\mathcal{G}}
\newcommand{\cV}{\mathcal{V}}

\newcommand{\cS}{\mathcal{S}}

\newcommand{\NM}[2]{\| #1 \|_{#2} }  

\usepackage{array}
\newcolumntype{L}[1]{>{\raggedright\let\newline\\\arraybackslash\hspace{0pt}}m{#1}}
\newcolumntype{C}[1]{>{\centering\let\newline  \\\arraybackslash\hspace{0pt}}m{#1}}
\newcolumntype{R}[1]{>{\raggedleft\let\newline \\\arraybackslash\hspace{0pt}}m{#1}}

\title{Hierarchical Heterogeneous Graph Representation Learning \\ for Short Text Classification}
 \author{Yaqing Wang$^{1}$, Song Wang$^{1,2}$, Quanming Yao$^{*3}$, Dejing Dou$^{1}$ \\
         $^{1}$Baidu Research, Baidu Inc., China 
         \\$^{2}$
         Department of ECE, University of Virginia, USA
         \\ $^{3}$
         Department of EE, Tsinghua University, China\\
     \texttt{\{wangyaqing01, v\_wangsong07, doudejing\}@baidu.com}\\
     \texttt{qyaoaa@tsinghua.edu.cn}}
     
\begin{document}
\maketitle

\begin{abstract}	
Short text classification is a fundamental task in natural language processing.  It is hard due to the lack of context information and labeled data in practice. In this paper, we propose a new method called SHINE, which is based on graph neural network (GNN), for short text classification.
First, we model the short text dataset as a hierarchical heterogeneous graph consisting of word-level component graphs which introduce more semantic and syntactic information. Then, we dynamically learn a short document graph that facilitates effective label propagation among similar short texts. Thus,
comparing with existing GNN-based methods, SHINE can better exploit interactions between
nodes of the same types and capture similarities between short texts. Extensive experiments on various benchmark short text datasets show that SHINE consistently outperforms state-of-the-art methods, especially with fewer labels.
\footnote{Codes are available at \url{https://github.com/tata1661/SHINE-EMNLP21}.
Correspondence author is Quanming Yao.}
\end{abstract}

\section{Introduction}
Short texts such as tweets, news feeds and web search snippets appear daily in our life~\cite{pang-lee-2005-seeing,phan2008learning}.  
To understand these short texts, 
short text classification (STC) is a fundamental task which can be found in many applications such as sentiment analysis~\cite{chen2019deep}, news classification~\cite{yao2019graph} and query intent classification~\cite{wang2017combining}.

STC is particularly hard in comparison to long text classification due to two key issues. 
The first key issue is that short texts only contain one or a few sentences whose overall length is small, 
which 
lack enough 
context information and strict syntactic structure to understand the meaning of texts~\cite{tang2015pte,wang2017combining}.  
For example, it is hard to get the meaning of  "\textit{Birthday girl is an amusing ride}" without knowing "\textit{Birthday girl}" is a 2001 movie. 
A harder case is to understand a web search snippet such as "\textit{how much Tesla}",  
which usually does not contain word order nor function words \cite{phan2008learning}. 
In addition, real STC tasks usually only have a limited number of labeled data compared to the  abundant unlabeled short texts emerging everyday~\cite{hu-etal-2019-heterogeneous}. 
Therefore, 
auxiliary knowledge is required to understand short texts, 
examples include concepts that can be found in common sense knowledge graphs~\cite{wang2017combining,chen2019deep}, 
latent topics extracted from the short text dataset~\cite{hu-etal-2019-heterogeneous}, 
and entities residing in knowledge graphs~\cite{hu-etal-2019-heterogeneous}. 
However, simply enriching auxiliary knowledge cannot solve the shortage of labeled data, which is another key issue commonly faced by real STC tasks~\cite{pang-lee-2005-seeing,phan2008learning}.  Yet the popularly used deep models require large-scale labeled data to train well~\cite{kim-2014-convolutional,liu2016recurrent}. 

Currently, graph neural networks (GNNs) designed for STC obtain the state-of-the-art  performance~\cite{hu-etal-2019-heterogeneous,ye2020document}. 
They both take the STC as the node classification problem on a graph with mixed nodes of different types:  
HGAT~\cite{hu-etal-2019-heterogeneous}  builds a corpus-level graph modeling latent topics, entities and documents and STGCN~\cite{ye2020document} operates on a corpus-level graph of latent topics, 
documents and words. In both works, 
each document is connected to its nodes of a different type such as entities and latent topics but not to other documents. 
However, 
they do not fully exploit interactions between nodes of the same type. 
They also fail 
to capture the similarities between short documents,
which is both useful to understand short texts~\cite{zhu2003semi,kenter2015short,wang2017combining} and 
and important to propagate few labels on graphs~\cite{kipf2016semi}.
Besides, both works have large parameter sizes: 
HGAT~\cite{hu-etal-2019-heterogeneous} is a GNN with dual-level attention and 
STGCN~\cite{ye2020document} merges the node representations with word embeddings obtained via a pretrained BERT~\cite{devlin-etal-2019-bert} via a bidirectional LSTM~\cite{liu2016recurrent}. 

To address the aforementioned  problems, 
we propose a novel HIerarchical heterogeNEous graph representation learning method for STC called SHINE, which is able to fully exploit interactions between
nodes of the same types and
capture similarity between short texts. 
SHINE operates on a hierarchically organized heterogeneous corpus-level graph, which consists of the following graphs at different levels: 
(i) {word-level component graphs} model interactions between words, part-of-speech (POS) tags and entities which can be easily extracted and carry additional semantic and syntactic information to compensate for the lack of context information; and  	
(ii) {short document graph} is dynamically learned and optimized to encode similarities between short documents which allows more effective label propagation among connected similar short documents. 
We conduct extensive experiments on a number of benchmark STC datasets including news, tweets, document titles and short reviews. 
Results show that the proposed SHINE consistently outperforms the state-of-the-art with a much smaller parameter size. 
\section{Related Works}
\label{sec:rel}

\subsection{Text Classification} 
\label{sec:texcls}

Text classification assigns predefined labels to documents of variable lengths which may consist of a single or multiple sentences~\cite{li2020survey}. 
Traditional methods adopt a two-step strategy: first extract human-designed features such as bag-of-words~\cite{blei2003latent} and term frequency-inverse document frequency (TF-IDF)~\cite{aggarwal2012survey} from documents, 
then learn classifiers such as support vector machine (SVM)~\cite{cortes1995support}.  
Deep neural networks such as convolutional neural networks (CNN)~\cite{kim-2014-convolutional} and long short-term memory (LSTM)
\cite{liu2016recurrent} can directly obtain expressive representations from raw texts and conduct classification in an end-to-end manner. 

Recently, 
graph neural networks (GNNs)~\cite{defferrard2016convolutional,kipf2016semi} 
have obtained the state-of-the-art performance on text classification.
They can be divided into two types. 
The first type of GNNs constructs document-level graphs where each document is modeled as a graph of word nodes,  
then formulates text classification as a whole graph classification problem~\cite{defferrard2016convolutional}. 
Examples are TLGNN~\cite{huang-etal-2019-text}, TextING~\cite{zhang-etal-2020-every}, HyperGAT~\cite{ding-etal-2020-less},
which establish word-word edges differently. 
In particular, some methods \cite{liu2019contextualized,chen2020iterative} propose to estimate the graph structure of the document-level graphs during learning. 
However,  
if only a few documents are labeled, these GNNs cannot work due to the lack of labeled graphs. 

As is known, GNNs such as graph convolutional network (GCN)~\cite{kipf2016semi} can conduct semi-supervised learning to solve node classification task on a graph where only a small number of nodes are labeled~\cite{kipf2016semi}.  
Therefore, another type of GNNs instead operates on a heterogeneous corpus-level graph which takes both text and word as nodes, 
and classifies unlabeled texts by node classification.     
Examples include TextGCN~\cite{yao2019graph}, TensorGCN~\cite{liu2020tensor}, HeteGCN~\cite{ragesh2021hetegcn} and TG-Transformer~\cite{zhang-zhang-2020-text} with different strategies to construct and handle heterogeneous nodes and edges.    
However, these methods cannot work well for short texts of limited length. 

\subsection{Short Text Classification (STC)} 
\label{sec:back_short_text} 

Short text classification (STC) is particularly challenging~\cite{aggarwal2012survey,li2020survey}. 
Due to limited length, short texts
lack context information and strict syntactic structure which are vital to text understanding~\cite{wang2017combining}.
Therefore, 
methods tailored for STC strive to incorporate various auxiliary information to enrich short text representations. 
Popularly used examples are concepts existing in external knowledge bases such as Probase~\cite{wang2017combining,chen2019deep} and latent topics discovered in the corpus~\cite{zeng-etal-2018-topic}. 
However, simply enriching semantic information cannot compensate for the shortage of labeled data, 
which is a common problem faced by real short texts such as queries and online reviews~\cite{pang-lee-2005-seeing,phan2008learning}. 
Thus, GNN-based methods which perform node classification for semi-supervised STC are utilized. 
HGAT~\cite{hu-etal-2019-heterogeneous} 
applies a GNN with dual-level attention to forward messages on a corpus-level graph modeling topics, entities and documents jointly, 
where the entities are words linked to knowledge graphs.
STGCN~\cite{ye2020document} operates on a corpus-level graph of topics, 
documents and words, and merges the node representations with word embeddings obtained via a pretrained BERT~\cite{devlin-etal-2019-bert} via a bidirectional LSTM~\cite{liu2016recurrent}. 
Currently,
the state-of-the-art method on 
STC is HGAT~\cite{hu-etal-2019-heterogeneous,yang2021hgat}.  

\begin{figure*}[htb]
	\centering
	\includegraphics[width = 1\textwidth]{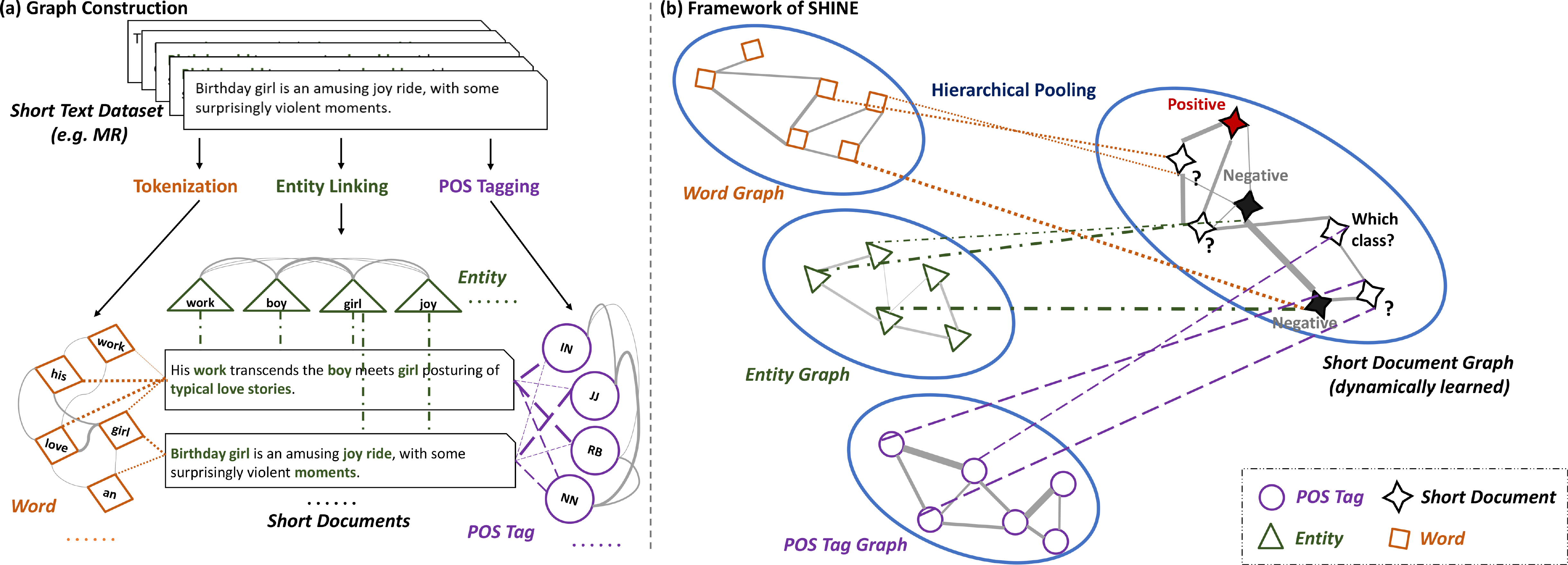}
	\caption{A high-level illustration of 
		(a) how we construct a heterogeneous corpus-level graph from a short text dataset using well-known natural language processing techniques; and  
		(b) framework of the proposed SHINE which hierarchically pools over word-level component graphs to obtain short document graph where node classification is conducted to classify those unlabeled nodes. 
		SHINE is trained end-to-end on the complete two-level graph with respect to the classification loss.
		The plotted examples of short texts are taken from the movie review (MR) dataset~\cite{pang-lee-2005-seeing}.}
	\label{fig:illus_heter}
		\vspace{-5px}
\end{figure*}

\section{Proposed Method}

As mentioned in Section~\ref{sec:back_short_text},
GNN-based methods,
i.e., HGAT and STGCN, 
can classify short texts
while HGAT performs better.
However, 
both works build a 
graph with mixed nodes of different types without fully exploiting interactions between nodes of the same type. 
Besides,
they fail 
to capture the similarities between short documents,
which can be important to propagate few labels on graphs.
Here, we present the proposed SHINE which 
can address above limitations,
thus
is able to better
compensate for the shortage of context information and labeled data
for STC.

Given a short text dataset\footnote{For consistency, we refer each input text to classify as a document which may consist of one or a few sentences following~\cite{tang2015pte,hu-etal-2019-heterogeneous}.} $\cS$ containing short documents, 
we model $\cS$ as a hierarchically organized heterogeneous graph consisting of:  
(i) word-level component graphs: we construct word-level component graphs which model word-level semantic and syntactic information in order to 
compensate for the lack of context information;
(ii) short document graph: 
we dynamically learn the short document graph via hierarchically pooling over word-level component graphs, 
such that the limited label information can be effectively propagated among similar short texts. 
A high-level illustration of SHINE is shown in 
Figure~\ref{fig:illus_heter}. 

In the sequel, vectors are denoted by lowercase
boldface, matrices by uppercase boldface. 
For a vector $\bx$, $[\bx]_i$ denotes the $i$th element of $\bx$. 
For a matrix $\bX$, 
$\bx^i$ denotes its $i$th row, 
$[\bX]_{ij}$ denotes the $(i,j)$th entry of $\bX$. 
For a set $\cS$, $|\cS|$ denotes the number of elements in $\cS$. 

\subsection{Word-Level Component Graphs}
\label{sec:wlcomp}

To compensate for the lack of 
context information and syntactic structure in short documents, 
we leverage various word-level components 
which can bring in more syntactic and semantic information. 
Particularly, we consider  the following three types of word-level components $\tau\in\{w,p,e\}$ in this paper: 
(i) word ($w$) which makes up short documents and carries semantic meaning;
(ii) POS tag ($p$)  which marks 
the syntactic role such as noun and verb of each word in the short text and is helpful for discriminating ambiguous words; and
(iii) entity ($e$) which corresponds to word that can be found in auxiliary knowledge bases such that  additional knowledge can be incorporated.  
SHINE can easily be extended with other components such as adding a topic graph on the first level. We use these three word-level components as they 
are well-known, easy to obtain at a low cost, and already surpass the state-of-the-art HGAT which use topics. 

We first provide a general strategy to obtain node embeddings from different types of word-level component graphs, 
then describe in detail how to construct  these graphs via common natural language processing techniques  including tokenization, entity linking and POS tagging.
In this way,
we can fully exploit interactions between nodes of the same type. 

\subsubsection{Node Embedding Learning}

Denote 
word-level component graph of type $\tau$ as 
$\cG_\tau=\{\cV_{\tau},\bA_{\tau}\}$ 
where $\cV_{\tau}$ is a set of nodes and $\bA_{\tau}\in\R^{|\cV_{\tau}|\times|\cV_{\tau}|}$ is the adjacency matrix.  
Each node $v^i_{\tau}\in\cV_{\tau}$ is provided with node feature $\bx^i_{\tau}\in\R^{d_{\tau}}$.  
For simplicity, the node features are collectively denoted as $\bX_{\tau}\in\R^{|\cV_{\tau}|\times d_{\tau}}$ with the $i$th row corresponds to one node feature $\bx^i_{\tau}$. 
These $\cG_\tau$s are used to capture the pairwise relationship between nodes of the same type, without being influenced by other types. 

Provided with $\cG_\tau$ and $\bX_{\tau}$, 
we use the classic 2-layer graph convolutional network (GCN)~\cite{kipf2016semi} to obtain node embeddings $\bH_\tau$. 
Formally, $\bH_\tau$ is updated as 
\begin{align}\label{eq:word-node}
	\bH_\tau= \tilde{\bA}_{\tau} \cdot \text{ReLu}(\tilde{\bA}_{\tau}\bX_{\tau}
	\bW_{\tau}^{1})\bW_{\tau}^{2},  
\end{align} 
where $[\text{ReLu}(\bx)]_i=\max([\bx]_i,0)$, ${\tilde{\bA}}_{\tau}=\bD_{\tau}^{-\frac{1}{2}}(\bI+\bA_{\tau})\bD_{\tau}^{-\frac{1}{2}}$ with ${[\bD_{\tau}]}_{ii}=\sum_j
[\bA_{\tau}]_{ij}$, and 
$\bW_{\tau}^{1},\bW_{\tau}^{2}$
are trainable parameters.

\subsubsection{Graph Construction}

Next, we present the details of how to construct each $\cG_{\tau}$ from $\cS$. 

\paragraph{Word Graph $\cG_w$.} 
We construct a word graph $\cG_w=\{\cV_w, \bA_w\}$ where word nodes are connected based on local co-occurrence relationships, while other types of relationship such as syntactic dependency~\cite{liu2020tensor} can also be used. 
We set 
$[\bA_w]_{ij}=\max(\text{PMI}(v_w^i, v_w^j),0)$ where PMI denotes the point-wise mutual information
between words $v_w^i, v_w^j\in \cV_w$~\cite{yao2019graph}.
We initialize the
node feature $\bx_w^i\in\R^{|\cV_w|}$ for $v_w^i\in\cV_w$ as a one-hot vector.  
Once learned by \eqref{eq:word-node}, $\bH_{w}$  is able to encode the topological structure of $\cG_w$ which is specific to $\cS$. 
We can also leverage generic semantic information by concatenating $\bH_{w}$ with pretrained word embeddings $\hat{\bH}_{w}$ extracted from large text corpus such as Wikidata~\cite{vrandevcic2014wikidata}. 

\paragraph{POS Tag Graph $\cG_p$.}  
We use the default POS tag set of NLTK\footnote{\url{http://www.nltk.org}} to obtain the POS tag for each word of short text in $\cS$, which forms the POS tag node set $\cV_p$.  
Similar to $\cG_w$, 
we construct a co-occurrence POS tag graph $\cG_p = \{\cV_p, \bA_p\}$ with $[\bA_p]_{ij}=\max(\text{PMI}(v_p^i, v_p^j),0)$, where the inputs are POS tags for all words. Then we again initialize the node feature $\bx_p^i\in\R^{|\cV_p|}$ as a one-hot vector.  

\paragraph{Entity Graph $\cG_e$.}
We obtain the entity node set $\cV_e$ by recognizing entities presented in the NELL knowledge base~\cite{carlson2010toward}.  
In contrast to words and POS tags which are abundant in the documents, the number of entities is much smaller. 
Most short documents only contain one entity, which makes it infeasible to calculate  co-occurrence statistics between entities. 
Instead, we first learn the
entity feature $\bx_e^i \in\R^{d_e}$ of each $v^i_e\in\cG_e$ from NELL, using the classic knowledge graph embedding method 
TransE~\cite{bordes2013translating}. 
Then, we measure the cosine similarity $c(v_e^i, v_e^j)$ between each entity pair $v_e^i, v_e^j\in \cV_e$ and set 
$[\bA_e]_{ij}=\max(c(x_e^j, x_e^j),0)$.

\subsection{Short Document Graph} 

As discussed in Section~\ref{sec:texcls},
the reason why GNN-based methods, 
which take short documents classification as node classification tasks,
can deal with few labels
is the usage of adjacent matrix which models the similarities between short documents.
However,
STGCN and HGAT do not 
consider such similarities.

Here,
to effectively propagate the limited label information, 
we dynamically learn the short document graph $\cG_s=\{\cV_s,\bA_s\}$ 
based on embeddings pooled over word-level component graph 
to encode the similarity between short documents, 
where $v^i_s\in\cV_s$ corresponds to one short document in $\cS$, 
and $\bA_s$ is the learned adjacency matrix. 
As shown in Figure~\ref{fig:illus_heter}, 
we propose to obtain $\cG_s$ via hierarchically pooling over word-level component graphs, 
hence $\cG_s$ is dynamically learned and optimized during training. 
This learned $\cG_s$ then facilitates efficient label propagation among connected short texts. 

\subsubsection{Hierarchical Pooling over $\cG_{\tau}$s}

In this section, 
we propose to learn $\bA_s$ via a text-specific hierarchically pooling over multiple word-level component graphs ($\cG_{\tau}$s). 

With $\bH_{\tau}$ obtained from \eqref{eq:word-node}, we represent each $v^i_s\in \cG_s$ 
by pooling over node embeddings in $\cG_{\tau}$ with respect to $\cG_\tau$: 
\begin{align}\label{eq:tau-to-text}
	\hat{\bx}^i_\tau=u(\bH^\top_{\tau} \bs^i_{\tau}), 	
\end{align}
where superscript $(\cdot)^\top$ denotes the transpose operation, and 
$u(\bx)=\bx/\NM{\bx}{2}$ normalizes $\bx$ to unit norm. 
Particularly, 
each $\bs^i_{\tau}$ is computed as follows: 
\begin{itemize}[leftmargin=*]
	\item When $\tau= w$ or $p$: $[\bs^i_{\tau}]_j=\text{TF-IDF}(v_{\tau}^j,v_s^i)$ where TF-IDF is the term frequency-inverse document frequency~\cite{aggarwal2012survey}.
	We then normalize $\bs^i_{\tau}$ as $\bs^i_{\tau}/\sum_{j=1}[\bs^i_{\tau}]_j$. 
	
	\item When $\tau=e$: 
	$[\bs^i_{e}]_j=1$ if $v_e^j$ exists in $v_s^i$ and $0$ otherwise, as most short texts only contain one entity.  $\bs^i_{e}$  is then normalized as $\bs^i_{e}/\sum_{j=1}[\bs^i_{e}]_j$. 
\end{itemize} 

This can be seen as explaining each short document from the perspective from words, POS tags and entities collectively. 
Finally, we obtain the short text representation $\bx^i_{s}$ of $v^i_s$ is 
obtained as 
\begin{align}\label{eq:text-feat}
	\bx^i_{s} = \hat{\bx}^i_w \mathbin\Vert \hat{\bx}^i_p \mathbin\Vert \hat{\bx}^i_e,  
\end{align}
where $\mathbin\Vert$ means concatenating vectors along the last dimension. 
Please note that concatenation is just an instantiation which already obtains good performance.  It can be replaced by more complex aggregation function such as weighted average or LSTM. 

\subsubsection{Dynamic Graph Learning}
\label{sec:dynamic-graph}

Now, 
we obtain $\bA_s$ on the fly using the learned short document features $\bx^i_s$'s: 
\begin{align}\label{eq:adj-short-text}
[\bA_s]_{ij} = 
\begin{cases}
(\bx^i_{s})^\top \bx^j_{s}
& \text{if\;} (\bx^i_{s})^\top \bx^j_{s} \ge \delta_s,
\\
0
& 
\text{otherwise}
\end{cases}, 
\end{align}
where  $\delta_s$ is a threshold used to sparsity $\bA_s$ such that 
short documents are connected only if they are similar enough viewed from the perspective of $\cG_{\tau}$s. 
Note that the resultant $\cG_s$ is dynamically changing along with the optimization process, where $\bH_{\tau}$,  $\bx^i_s$, and $\bA_s$ are all optimized and improved.

Upon this $\cG_s$, we propagate label information among similar short documents via a 2-layer GCN.  
Let $\bX_s$ collectively record short text embeddings with $\bx^i_s$ on the $i$th row. 
The class predictions of all short documents in $\cS$ with respect to $C$ classes are obtained as
\begin{align}\label{eq:predict}
\!\hat{\bY}_s\!=\!\text{softmax}
\big(
{\bA}_{s}
\!
\cdot
\!
\text{ReLu}({\bA}_{s} \bX_{s} \bW_{s}^1) 
\!
\cdot
\! 
\bW_{s}^2
\big),\!  
\end{align} 
where $[\text{softmax}(\bx)]_i=\exp([\bx]_i)/\sum_j\exp([\bx]_j)$ is applied for each row, 
$\bW_{s}^1$ and $\bW_{s}^2$ are trainable parameters. 

We train the complete model by optimizing the cross-entropy loss function in an end-to-end manner: 
\begin{align}\label{eq:loss}
\mathcal{L}=-\sum\nolimits_{i\in\mathcal{I}_l}(\by^i_s)^\top\log(\hat{\by}^i_s), 
\end{align}
where $\mathcal{I}_l$ records the 
indices of the labeled short documents, 
$\by^i_s\in\R^C$ is a one-hot vector with all 0s but a single one denoting the index of the ground truth class $c\in\{1,\dots,C\}$. 
By jointly optimized with respect to the single objective, 
different types of graphs can influence each other.
During learning, node embeddings of $\cG_\tau$ for all $\tau\in\{w,p,e,s\}$ and $\bA_{s}$ are all updated. 
The complete procedure of SHINE is shown in Algorithm~\ref{alg:SHINE}.

\begin{algorithm}[ht]
\caption{SHINE Algorithm.}
\begin{algorithmic}[1]
	\REQUIRE short text dataset $\cS$, word-level component graphs $\cG_{\tau}=\{\cV_{\tau},\bA_{\tau}\}$ with node features $\bX_{\tau}$, 
	sample-specific aggregation vectors $\{ \bs^i_{\tau} \}$ where $\tau\in\{w,p,e\}$; 
	\FOR{$t = 1,2,\dots T$}
	\FOR{$\tau\in\{w,p,e\}$}
	\STATE obtain node embeddings $\bH_{\tau}$ of $\cG_{\tau}$ by \eqref{eq:word-node}; 
	\ENDFOR
	\STATE obtain short document features $\bX_s$ via hierarchically pooling over $\cG_{\tau}$s by \eqref{eq:text-feat};
	\STATE obtain short document embeddings from $\cG_s$ and make the class prediction by \eqref{eq:predict};
	\STATE optimize model parameter with respect to \eqref{eq:loss} by back propagation;
	\ENDFOR
\end{algorithmic}
\label{alg:SHINE}
\end{algorithm}

\section{Experiments}

All results are averaged over five runs and are obtained on a PC with 32GB memory, Intel-i8 CPU, and a 32GB NVIDIA Tesla V100 GPU.
\begin{table*}[htbp]
	\centering
	\setlength\tabcolsep{3pt}
	\begin{tabular}
		{c | c | c | c|c | c | c|c}
		\hline
		&	\# texts    & avg. length    &  \# classes &\# train (ratio)&\# words&\# entities &\# POS tags \\ \hline
		{Ohsumed}&7,400&6.8&23&460 (6.22\%)&11,764&4,507&38\\
		{Twitter}&10,000&3.5&2&40 (0.40\%)&21,065&5,837&41\\
		{MR}&10,662&7.6&2&40 (0.38\%)&18,764&6,415&41\\
		{Snippets}&12,340&14.5&8&160 (1.30\%)&29040&9737&34\\
		{TagMyNews}&32,549&5.1&7&140 (0.43\%)&38629&14734&42\\\hline
	\end{tabular}
		\vspace{-5px}
	\caption{Summary of short text datasets used.}
	\label{tab:data-stat}
		\vspace{-5px}
\end{table*}
\subsection{Datasets} 

We perform experiments on a variety of 
publicly accessible benchmark short text datasets (Table~\ref{tab:data-stat}):  
\begin{enumerate}[label=(\roman*)] 
\item \textbf{Ohsumed}\footnote{\url{http://disi.unitn.it/moschitti/corpora.htm}}: a subset of the bibliographic Ohsumed dataset~\cite{hersh1994ohsumed} used in~\cite{hu-etal-2019-heterogeneous} where the title is taken as the short text to classify. 
\item \textbf{Twitter}\footnote{\url{http://www.nltk.org/howto/twitter.html\#corpus_reader}}: a collection of tweets expressing positive or negative attitude towards some contents. 
\item \textbf{MR}\footnote{\url{http://www.cs.cornell.edu/people/pabo/movie-review-data/}}: a movie review dataset for sentiment analysis~\cite{pang-lee-2005-seeing}.  
\item \textbf{Snippets}\footnote{Snippets and TagMyNews are downloaded from \url{http://acube.di.unipi.it:80/tmn-dataset/}.}: a  dataset of web search snippets returned by Google Search~\cite{phan2008learning}.
\item \textbf{TagMyNews}: a dataset contains English news titles collected from Really Simple Syndication (RSS) feeds, as adopted by~\citet{hu-etal-2019-heterogeneous}. 
\end{enumerate}

We tokenize each sentence and remove stopping words and low-frequency words which appear less than five times in the corpus as suggested in~\cite{yao2019graph,hu-etal-2019-heterogeneous}. 

Following~\cite{hu-etal-2019-heterogeneous}, we randomly sample 40 labeled short documents from each class where 
half of them forms 
the training set and the other half forms the validation set for hyperparameter tuning. 
The rest short documents are taken as the test set, which are unlabeled during training.  

\begin{table*}[ht]
\centering
\setlength\tabcolsep{5pt}
\begin{tabular}{c|c|cc|cc|cc|cc|cc}
	\hline
	&           & \multicolumn{2}{c|}{Ohsumed} & \multicolumn{2}{c|}{Twitter} & \multicolumn{2}{c|}{MR} & \multicolumn{2}{c|}{Snippets} & \multicolumn{2}{c}{$\!\!$TagMyNews$\!\!$} \\ \hline
	{$\!\!$Group$\!\!$} &  {Model}  &      ACC       &     F1      &      ACC       &     F1      &      ACC       &   F1   &  ACC  &          F1           &      ACC       &            F1            \\ \hline
	(A)   & TFIDF+SVM &     39.02      &    24.78    &    53.69     &    52.45    &    54.29      & 48.13  & 64.70 &         59.17         &    39.91      &          32.05           \\
	&  LDA+SVM  &    38.61      &    25.03    &     54.34      &    53.97    &    54.40     & 48.39  & 62.54 &         56.40         &     40.40      &          30.40           \\
	&    PTE    &     36.63      &    19.24    &     54.24      &    53.17    &     54.74      & 52.36  & 63.10 &         58.96         &     40.32      &          33.56           \\\hline
	(B)     & BERT-avg  &     23.91      &4.98             &     54.92      & 51.15            &     51.69      &  50.65      & 79.31 &        78.47               &     55.13      &   44.26                       \\
	& BERT-CLS  &     21.76      &    4.81     &     52.00      &    43.34    &     53.48      & 46.99  & 81.53 &         \underline{79.03}         &     58.17      &          41.04           \\ \hline
	(C)   & CNN-rand  &     35.25      &    13.95    &     52.58      &    51.91    &     54.85      & 51.23  & 48.34 &         42.12         &     28.76      &          15.82           \\
	&  CNN-pre  &     32.92      &    12.06    &     56.34      &    55.86    &     58.32      & 57.99  & 77.09 &         69.28         &     57.12      &          45.37           \\
	& LSTM-rand &     23.30      &    5.20     &     54.81      &    53.85    &     53.13      & 52.98  & 30.74 &         25.04         &     25.89     &          17.01           \\
	& LSTM-pre  &     29.05      &    5.09     &    58.20      &    58.16    &     59.73      & 59.19  & 75.07 &         67.31             &     53.96      &    42.14                   \\\cmidrule{2-12}
	&   TLGNN   &     35.76      &   13.12          &    58.33      & 53.86            &   58.48             & 58.45       &  70.25     & 63.18                      & 44.43               &    32.33                      \\
	&  TextING  &     38.27      & 21.34            &     59.79      &  59.44           &     58.89      &    58.76    & 71.13 & 70.71                      & 52.53 & 40.20                         \\
	& HyperGAT  &     36.60      & 19.98           &     58.42      &     53.71        &     58.65      & 58.62       & 70.89 &   63.42                    &     45.60      &  31.51                        \\\cmidrule{2-12}
	&  TextGCN  &     41.56      &   \underline{ 27.43}    &     60.15      &    59.82    &     59.12      & 58.98  & 77.82 &         71.95         &     54.28      &          46.01           \\		        		        
	& TensorGCN &    41.84      & 24.24             &    61.24      &   61.19           &     59.22      & 58.78       & 74.38 &   73.96                    &     55.58      &   43.21                       \\\hline
	(D)    &   STCKA   &     30.19      & 10.12            &     57.45      &   56.97          &     53.22      &    50.11    & 68.96 &   61.27                    &     30.44      & 20.01                         \\ \cmidrule{2-12}
	&   HGAT    &     \underline{42.68}      &    24.82    &     63.21      &    62.48    &     \underline{62.75}      & \underline{62.36}  & \underline{82.36} &         74.44         &     \underline{61.72}      &          \underline{53.81}           \\
	&   STGCN   &      33.91          &   27.22          &     \underline{64.33}           &    \underline{64.29}         &      58.18          &    58.11    &  70.01     &     69.93                  &     34.74           &34.01                          \\
	&   SHINE (ours)   & \textbf{45.57} &    \textbf{30.98}         & \textbf{72.54} &  \textbf{72.19}           & \textbf{64.58} &  \textbf{63.89}      &    \textbf{82.39}   &         \textbf{81.62}              &           \textbf{62.50}           & \textbf{56.21}                        \\ \hline
& relative $\uparrow$ (\%) & 6.77&   12.94      & 12.76&  12.29    & 2.92 & 2.45    &  0.85&       3.17&   1.26   &  4.46                        \\ \hline
\end{tabular}
	\vspace{-5px}
	\caption{Test performance (\%) measured on short text datasets. 		
		The best results (according to the pairwise t-test with 95\% confidence) are highlighted in bold. The second best results are marked in Italic.
		The last row records the relative improvement (\%) of SHINE over the second best result. 
	}
\label{tab:results}
	\vspace{-10px}
\end{table*}

\subsection{Compared Methods} 

The proposed \textbf{SHINE} is compared with the following methods.  
\begin{itemize}[leftmargin=*]
\setlength{\itemsep}{0pt}
\setlength{\parskip}{0pt}
\item \textit{Group (A).}
Two-step feature extraction and classification methods include
(i) \textbf{TF-IDF+SVM} and (ii) \textbf{LDA+SVM}\cite{cortes1995support} which use support vector machine to classify documents represented by TF-IDF feature and LDA feature respectively; and 
(iii) \textbf{PTE}\footnote{\url{https://github.com/mnqu/PTE}}~\cite{tang2015pte} which learns a linear classifier upon documents represented as the average word embeddings pretrained from bipartite word-word, word-document and word-label graphs. 

\item \textit{Group (B).} 
\textbf{BERT}\footnote{\url{https://tfhub.dev/tensorflow/bert_en_uncased_L-12_H-768_A-12/4}}~\cite{devlin-etal-2019-bert} which is pretrained on a large corpus and fine-tuned together with a linear classifier for the short text classification task. Each document is represented as 
the averaged word embeddings (denote as \textbf{-avg}) or the embedding of the CLS  token (denote as \textbf{-CLS}). 

\item \textit{Group (C).} 
Deep text classification methods include 
(i) \textbf{CNN}~\cite{kim-2014-convolutional} and 
(ii) \textbf{LSTM}~\cite{liu2016recurrent} where the input word embeddings are either randomly initialized (denote as \textbf{-rand}) or pretrained from large text corpus (denote as \textbf{-pre}); 
GNNs which perform graph classification on document-level graphs including  
(iii) \textbf{TLGNN}\footnote{\url{https://github.com/LindgeW/TextLevelGNN}} 
\cite{huang-etal-2019-text},  
(iv) \textbf{TextING}\footnote{\url{https://github.com/CRIPAC-DIG/TextING}}
\cite{zhang-etal-2020-every}, and 
(v) \textbf{HyperGAT}\footnote{\url{https://github.com/kaize0409/HyperGAT}} 
\cite{ding-etal-2020-less}; 
GNNs which perform node classification on corpus-level graphs including
(vi) \textbf{TextGCN}\footnote{\url{https://github.com/yao8839836/text_gcn}} 
\cite{yao2019graph} and  
(vii) \textbf{TensorGCN}\footnote{\url{https://github.com/THUMLP/TensorGCN}} 
\cite{liu2020tensor}. 
HeteGCN~\cite{ragesh2021hetegcn} and TG-Transformer~\cite{zhang-zhang-2020-text} are not compared due to the lack of publicly available codes. 	

\item \textit{Group (D).} Deep STC methods including 
(i) \textbf{STCKA}\footnote{\url{https://github.com/AIRobotZhang/STCKA}}
\cite{chen2019deep}: 
an attention-based BiLSTM model, 
which fuses concept found in auxiliary  knowledge bases into short document embedding; 
(ii) \textbf{HGAT}\footnote{\url{https://github.com/ytc272098215/HGAT}}  
	\cite{hu-etal-2019-heterogeneous} which operates on a corpus-level graph of entities, topics and documents using a GNN with dual-level attention; 
	and (iii) \textbf{STGCN}\footnote{\url{https://github.com/yzhihao/STGCN}}~\cite{ye2020document} which operates on a corpus-level graph of words, topics and documents and uses a bidirectional LSTM to merge the word embeddings learned by a GNN with 
	word embeddings produced by a pretrained BERT. 

\end{itemize} 
For these baseline methods, we either show the results reported in previous research \cite{hu-etal-2019-heterogeneous,yang2021hgat} or run the public codes provided by the authors. 
For fairness, we use the 
public 300-dimensional GloVe word embeddings\footnote{\url{http://nlp.stanford.edu/data/glove.6B.zip}} in all methods which require pretrained word embeddings~\cite{pennington-etal-2014-glove}. 

\paragraph{Hyperparameter Setting.}
For all methods, we find hyperparameters using the validation set via grid search.  
For SHINE, we set entity embedding dimension $d_e$ as 100.
For all the datasets, we set the sliding window size of PMI as 5 for both $\cG_w$ and $\cG_p$,    
set the embedding size of all GCN layers used in SHINE as 200, and set the threshold 
$\delta_s$ for $\cG_s$ as 2.5.   
We implement SHINE in PyTorch and train the model for a maximum number of 1000 epochs using Adam~\cite{kingma2014adam} with learning rate $10^{-3}$.   
We early stop training if the validation loss does not decrease for 10 consecutive epochs. 
Dropout rate is set as 0.5. 

\paragraph{Evaluation Metrics.}
We evaluate the classification performance using test accuracy (denote as ACC in short)
and macro-averaged F1 score (denote as F1 in short) following~\cite{tang2015pte,yang2021hgat}. 

\subsection{Benchmark Comparison}

\paragraph{Performance Comparison.}
Table~\ref{tab:results} shows the performance. 
As can be seen, GNN-based methods in group (D) obtain better classification results in general, 
where the proposed SHINE consistently obtains the state-of-the-art test accuracy and macro-F1 score. 
This can be attributed to the effective semantic and syntactic information fusion and the modeling of short document graphs.

In addition, 
if we order datasets by increasing average text length (i.e., Twitter, TagMyNews, Ohsumed, MR and Snippets), we can find that 
SHINE basically obtains larger relative improvement over the second best method on shorter documents as shown in the last row of Table~\ref{tab:results}.  
This validates the efficacy of label propagation in SHINE, which can be attributed to the dynamical learning of short document graph. 
As shown, GNNs which perform node classification on the corpus-level graph obtain better performance than 
GNNs which perform graph classification on short text datasets with a few labeled short documents. 
Another common observation is that incorporating pretrained word embeddings can consistently improve the accuracy, as can be observed by comparing CNN-pre to CNN-rand, LSTM-pre to LSTM-rand, BiLSTM-pre to BiLSTM-rand. 
CNN and LSTM can obtain worse performance than traditional methods in group (A), such as results on Ohsumed.  
The fine-tuned BERT encodes generic semantic information from a large corpus, but it cannot beat SHINE which is  particularly designed to handle the short text dataset.

\paragraph{Model Size Comparison.}
Table~\ref{tab:mem} presents the parameter size of SHINE and the two most relevant GNN-based methods,
i.e., HGAT and STGCN. 
As can be seen, SHINE takes much smaller parameter size. 
The reason is that
instead of organizing different types of nodes in the same graph like HGAT and STGCN,
SHINE separately constructs graphs for each
type of nodes
and pools from them to represent short documents.
Thus,
the graph used in SHINE can be much smaller than 
HGAT and STGCN, which leads to a reduction of the parameter number. 
We also observe that SHINE takes less training time per epoch. 

\begin{table}[htbp]
	\centering
	\setlength\tabcolsep{5pt}
	\begin{tabular}{c | c | c | c}
		\hline
		&    HGAT    & STGCN    &  SHINE \\ \hline
	{Ohsumed}&3,091,523&2,717,104&212,146\\
	{Twitter}&2,312,672&3,201,824&201,604\\
	{MR}&6,030,640&3,326,224&201,604\\
	{Snippets}&8,892,778&3,238,304&204,616\\
	{$\!\!$TagMyNews$\!\!$}&10,162,899&6,653,024&204,114\\\hline
	\end{tabular}
	\vspace{-5px}
	\caption{Comparison of model parameter size.}
	\label{tab:mem}
		\vspace{-10px}
\end{table}

\subsection{Ablation Study}
\begin{figure}[htb]
	\centering
	\includegraphics[width = 0.32\textwidth]{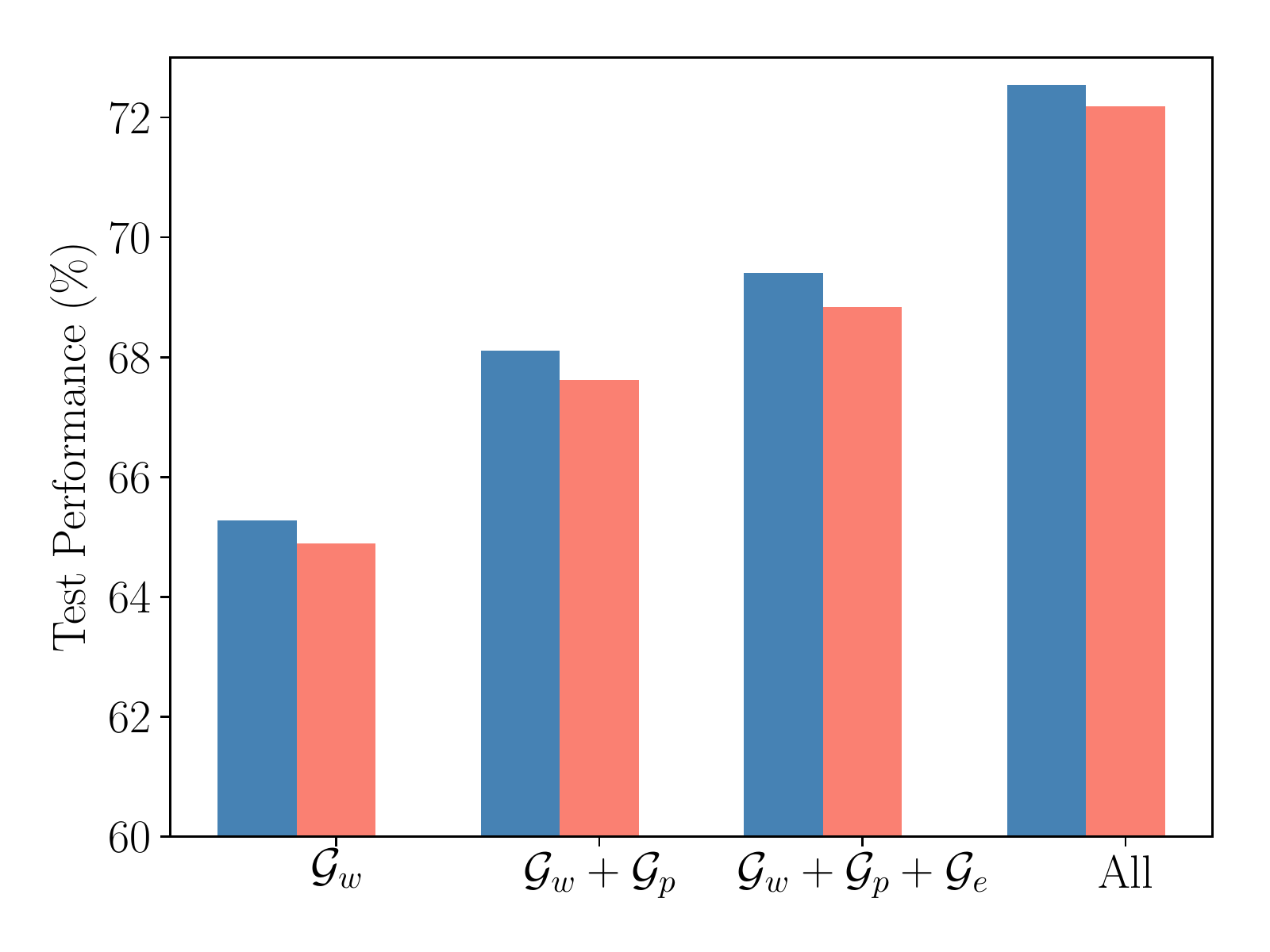}
	
		\vspace{-5px}
	\caption{Effect of adding each graph component.}
	\label{fig:abla-component}
\end{figure}

We compare with 
different variants of
SHINE 
to evaluate the contribution of each part:
\begin{enumerate}[label=(\roman*)] 
\item \textbf{w/o $\cG_w$}, \textbf{w/o $\cG_p$} and \textbf{w/o $\cG_e$}: remove one single $\cG_\tau$ from SHINE while keeping the other parts unchanged.
\item \textbf{w/o pre}: do not concatenate $\bH_w$ with pretrained word embeddings $\hat{\bH}_{w}$. 
\item \textbf{w/ pre $\bX_w$}: initializes node embeddings $\bX_w$ of $\cG_w$ as pretrained word embeddings $\hat{\bH}_{w}$ directly. 
\item \textbf{w/o word GNN}:  
fix $\bH_\tau$ as the input node features $\bX_\tau$ of each $cGH_\tau$, therefore the node embeddings of $\cG_s$ are simply weighted average of corresponding word-level features.
\item \textbf{w/o doc GNN}: use label propagation \cite{zhou2004learning} to directly obtain class prediction using $\bA_s$ learned by \eqref{eq:adj-short-text} and $\bx^i_s$ learned by \eqref{eq:text-feat}. 
\item \textbf{w/ a single GNN}: run a single GNN on a heterogeneous corpus-level graph containing the same set of words, entities, POS tags and documents as ours. We modify TextGCN \cite{yao2019graph} to handle this case.
\end{enumerate}
Table~\ref{tab:results-abla} shows the results. 
As shown, word-level component graphs and short document graph contribute larger to the effectiveness of SHINE. 
The concatenation of pretrained word embedding can slightly improve the performance. 
However, 
``w/ pre $\bX_w$" is worse than SHINE. 
This shows the benefits of separating corpus-specific and general semantic information: using $\cG_w$ with one-hot initialized features to capture corpus-specific topology among words, while using pretrained word embeddings to bring in general semantic information extracted from an external large corpus.  
The performance gain of SHINE with respect to ``w/o word GNN" validates the necessity of (i) message passing among nodes of the same type and update node embeddings accordingly and (ii) update $\cG_\tau$s with respect to the STC task. 
While the improvement of SHINE upon ``w/o doc GNN" shows that refining short document embeddings by GNN is useful. 
Finally, SHINE defeats ``w/ a single GNN" which  uses the same amount of information.   
This reveals that SHINE outperforms due to model design.
Figure~\ref{fig:abla-component} further plots the influences of incrementally adding in more word-level component graphs and the short document graph. 
This again validates the effectiveness of SHINE framework. 

\begin{table*}[ht]
	\centering

	\begin{tabular}{c|cc|cc|cc|cc|cc}
		\hline
		&  \multicolumn{2}{c|}{Ohsumed}   &  \multicolumn{2}{c|}{Twitter}   &     \multicolumn{2}{c|}{MR}     &  \multicolumn{2}{c|}{Snippets}  & \multicolumn{2}{c}{$\!\!$TagMyNews$\!\!$} \\ \hline
		{Model}   &      ACC       &       F1       &      ACC       &       F1       &      ACC       &       F1       &      ACC       &       F1       &      ACC       &            F1            \\ \hline		 
		w/o $\cG_w$ &     21.91      &     11.87      &     60.93      &     60.39      &     55.03      &     54.00      &     70.25      &     68.82      &     55.68      &          48.38           \\
		w/o $\cG_p$ &     26.89      &     13.10      &     67.37      &     66.78      &     60.21      &     59.79      &     77.66      &     75.86      &     60.37      &          52.96           \\
		w/o $\cG_e$ &     30.17      &     15.31      &     68.46      &     67.89      &     61.54      &     60.60      &     80.48      &     77.82      &     60.44      &          54.10           \\
		w/o $\cG_s$ &     33.20      &     18.93      &     68.53      &     68.19      &     61.08      &     60.87      &     78.68      &     77.74      &     61.05      &          54.51           \\
		w/o pre   &     36.23      &     21.50      &     69.04      &     68.57      &     61.97      &     61.31      &     78.47      &     78.01      &     61.09      &          54.57           \\
		w/ pre $\bX_w$& 27.13&19.94&66.70&66.35&60.56&60.51&71.70&70.52&57.04&50.11\\
		w/o word GNN& 25.60&17.14&54.82&53.85&53.49&52.74&65.43&64.62&54.96&45.32\\
		w/o doc GNN&37.41&25.85&70.60&70.47&61.94&61.60&79.27&78.10&61.38&55.93\\
		w/ a single GNN& 42.56&28.18&61.35&61.20&60.39&60.21&78.52&73.64&56.58&48.18\\
		SHINE    & \textbf{45.57} & \textbf{30.98} & \textbf{72.54} & \textbf{72.19} & \textbf{64.58} & \textbf{63.89} & \textbf{82.39} & \textbf{81.62} & \textbf{62.50} &      \textbf{56.21}  
		\\ \hline
	\end{tabular}
	\vspace{-5px}
	\caption{Test performance (\%) of SHINE and its variants on short text datasets.         
		The best results (according to the pairwise t-test with 95\% confidence) are highlighted in bold.}
	\label{tab:results-abla}
	\vspace{-10px}
\end{table*}

\begin{figure*}[t]
	\centering
	\subfigure[Varying labled data proportion (\%).\label{fig:abla-ratio}]
	{
		\includegraphics[width =
		0.32\textwidth]{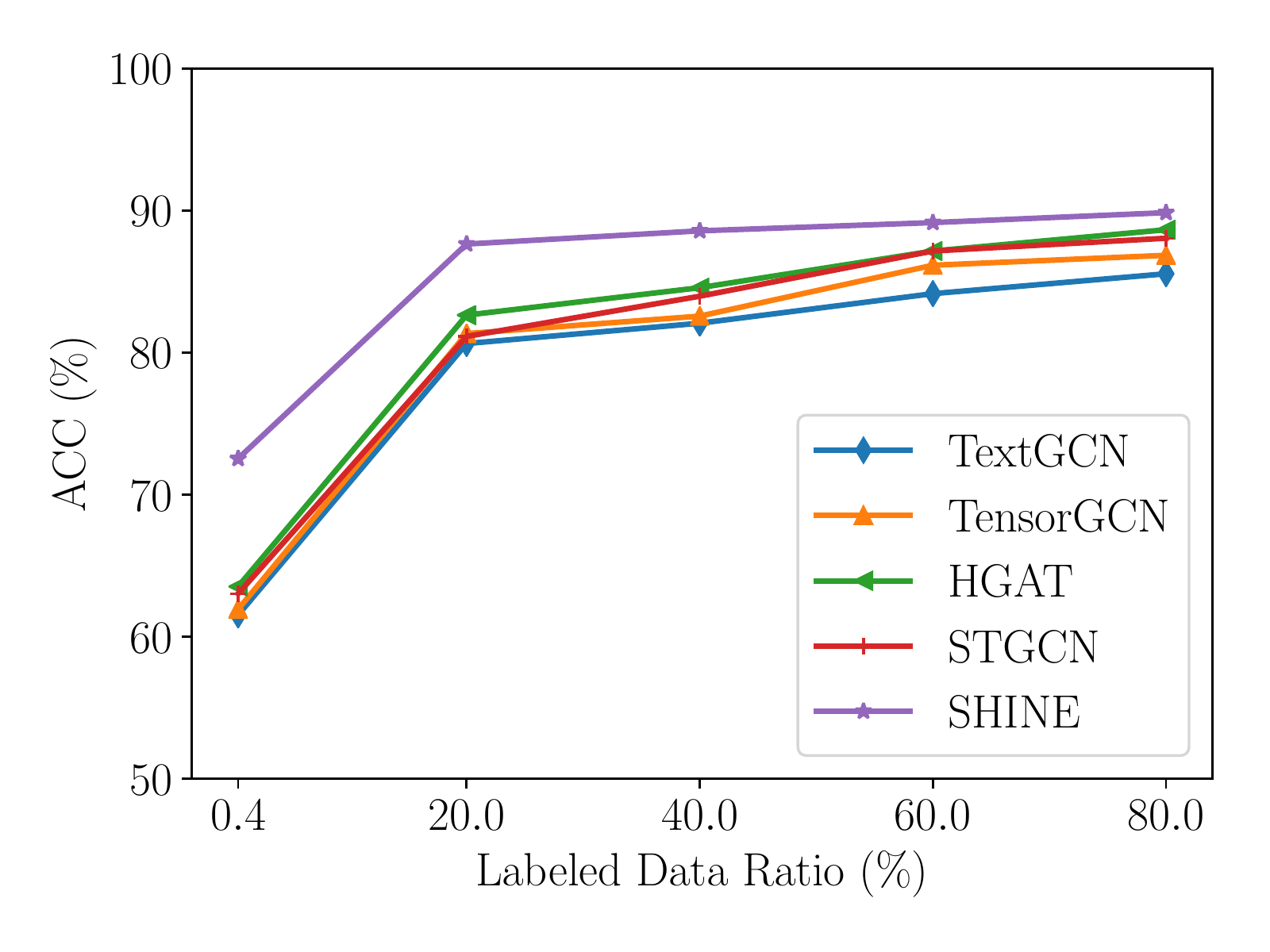}}
	\subfigure[Varying $\delta_s$ in \eqref{eq:adj-short-text}. \label{fig:abla-threshold}]{
		\includegraphics[width =
		0.32\textwidth]{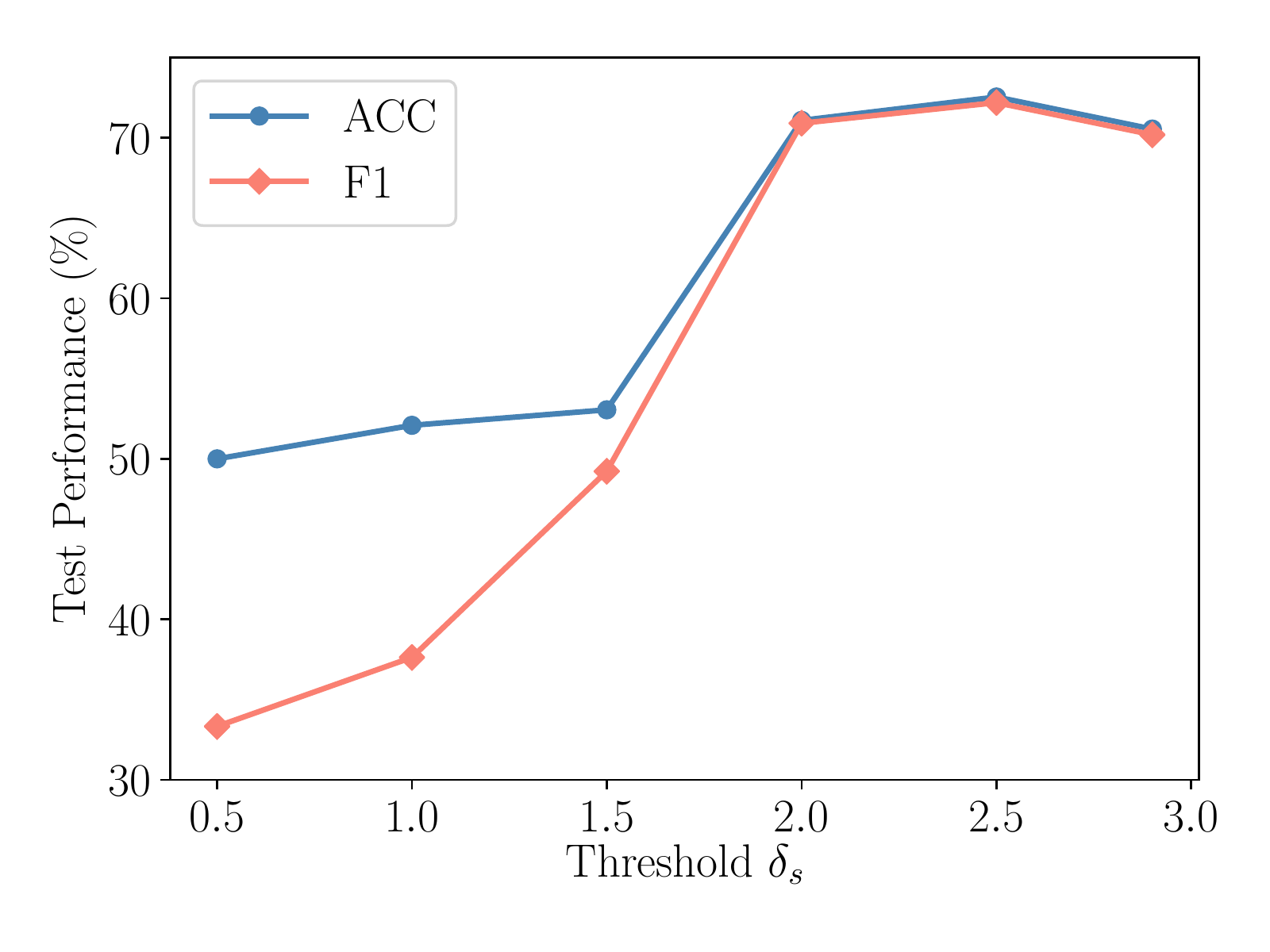}}
	\subfigure[Varying the embedding size of GCN.\label{fig:abla-embedding}]{
		\includegraphics[width =
		0.32\textwidth]{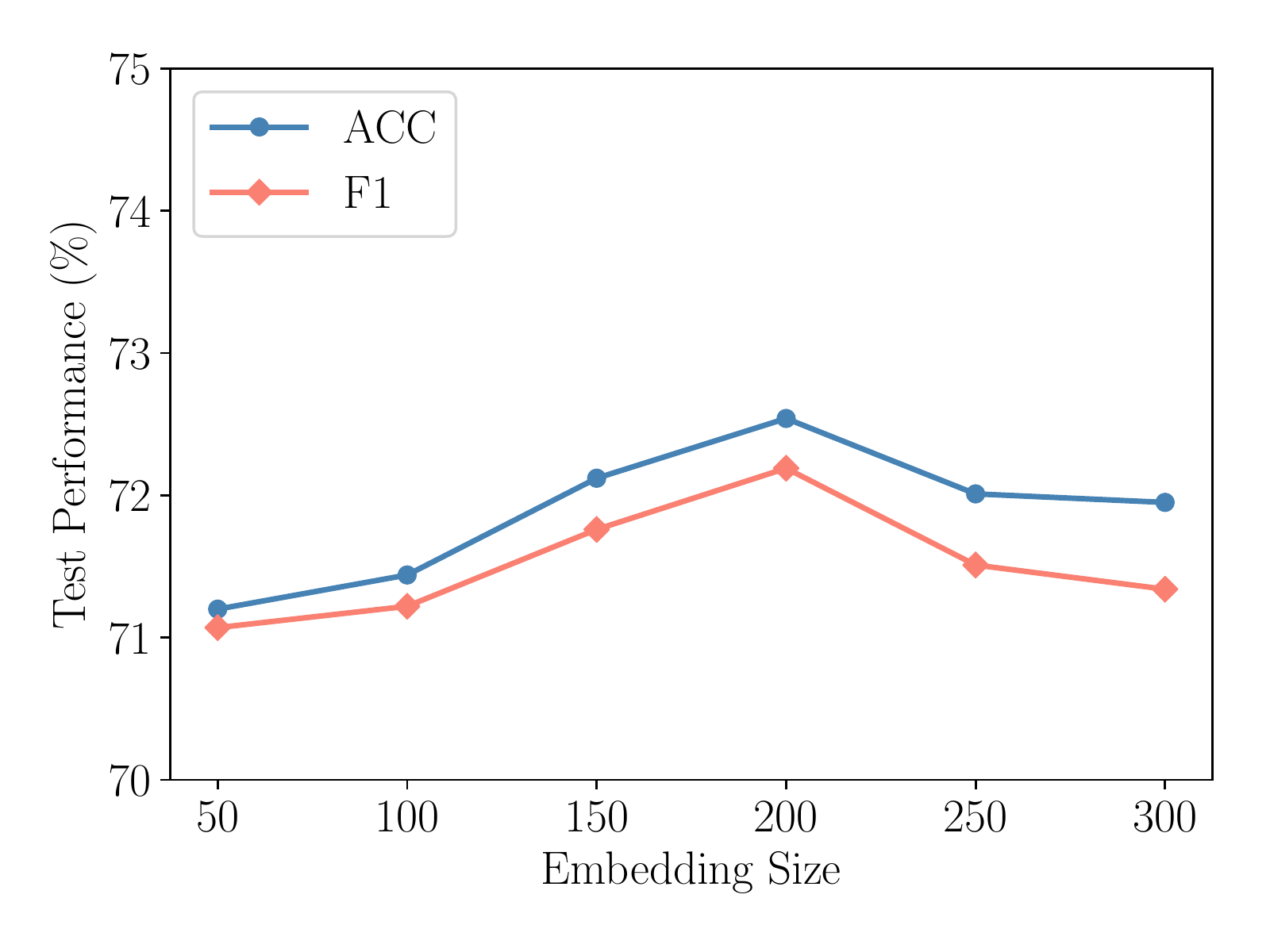}}
	
	\vspace{-5px}
	\caption{Model sensitivity analysis of SHINE on Twitter.}
	\label{fig:abla}
		\vspace{-10px}
\end{figure*}

\subsection{Model Sensitivity} 
  
We further examine
the impact of labeled training data proportion for GNN-based methods which perform node classification on the corpus-level graph, including TextGCN, TensorGCN, HGAT, STGCN and SHINE. 
Figure~\ref{fig:abla-ratio} plots the results. 
As shown, SHINE consistently outperforms other methods, where the performance gap is increasing with fewer labeled training data . 
Figure~\ref{fig:abla-threshold} plots the impact of threshold $\delta_s$ in \eqref{eq:adj-short-text}. 
At first, performance increases with a larger $\delta_s$ which 
leads to a sparse $\cG_s$ where only certainly similar short documents are connected to propagate information. However, when $\delta_s$ is too large, $\cG_s$ losses its functionality and reduces to w/o $\cG_{\tau}$ in Table~\ref{tab:results-abla}. 
Finally, recall that we set the embedding size of all GCN layers used in SHINE equally. 
Figure~\ref{fig:abla-embedding} plots the effect of varying this embedding size. As observed, small embedding size cannot capture enough information while a overly large embedding size may not improve the performance but is more computational costly. 

\newpage

\section{Conclusion}

In this paper, we propose SHINE, 
a novel hierarchical heterogeneous graph representation learning method for short text classification. 
It is particularly useful to compensate for the lack of context information and propagate the limited number of labels efficiently. 
Specially, SHINE can effectively learn from a hierarchical graph modeling different perspectives of the short text dataset: 
word-level component graphs are used to understand short texts from the semantic and syntactic perspectives,  
and the dynamically learned short document graph allows efficient and effective label propagation among similar short documents. 
Extensive experiments show that SHINE outperforms the others consistently. 

As for the future works,
we plan to search graph structure~\cite{zhao2021search}
and utilize automated machine learning~\cite{yao2018taking} 
to 
improve learning performance.

\clearpage
\newpage
{

\section*{Acknowledgements}
We thank the anonymous reviewers for their valuable comments. 
Parts of experiments were carried out on Baidu Data Federation Platform. 
Correspondence author is Quanming Yao.
\bibliography{hin}
\bibliographystyle{acl_natbib}
}

\appendix

\end{document}